# *Mind the Pause*: Disfluency-Aware Objective Tuning for Multilingual Speech Correction with LLMs

**Deepak Kumar, Baban Gain & Asif Ekbal**
Department of Computer Science and Engineering
Indian Institute of Technology Patna
Patna, India-801106
deepakkumar1538@gmail.com, {baban_2321cs12, asif}@iitp.ac.in

## Abstract

Automatic Speech Recognition (ASR) transcripts often contain disfluencies, such as fillers, repetitions, and false starts, which reduce readability and hinder downstream applications like chatbots and voice assistants. If left unaddressed, such disfluencies can significantly degrade the reliability of downstream systems. Most existing approaches rely on classical models that focus on identifying disfluent tokens for removal. While this strategy is effective to some extent, it often disrupts grammatical structure and semantic coherence, leading to incomplete or unnatural sentences. Recent literature explored the use of large language models (LLMs); however, these efforts have primarily focused on disfluency detection or data augmentation, rather than performing comprehensive correction. We propose a multilingual correction pipeline where a sequence tagger first marks disfluent tokens, and these signals guide instruction fine-tuning of an LLM to rewrite transcripts into fluent text. To further improve reliability, we add a contrastive learning objective that penalizes the reproduction of disfluent tokens, encouraging the model to preserve grammar and meaning while removing disfluent artifacts. Our experiments across three Indian languages, namely Hindi, Bengali, and Marathi show consistent improvements over strong baselines, including multilingual sequence-to-sequence models. These results highlight that detection-only strategies are insufficient. Combining token-level cues with instruction tuning and contrastive learning provides a practical and scalable solution for multilingual disfluency correction in speech-driven NLP systems. We make the codes publicly available at https://github.com/deepak-kumar-98/Mind-the-Pause.

## 1 Introduction

Spontaneous speech is rarely fluent: speakers often produce *disfluencies*, such as fillers (e.g., *uh, um*), repetitions, false starts, and discourse markers. These disfluencies are often preserved by Automatic Speech Recognition (ASR) systems, degrading the readability of transcripts and reducing the reliability of downstream natural language processing (NLP) applications, such as chatbots, task-oriented dialogue agents, and voice assistants. When propagated into downstream tasks, disfluent tokens cause misinterpretations, incoherent responses, and overall poor user experience.

Most of the existing works treat disfluency handling as a detection-then-deletion problem: models tag disfluent tokens and then remove them from transcripts (Johnson and Charniak, 2004; Rasooli and Tetreault, 2013; Honnibal and Johnson, 2014; Zayats et al., 2016; Hough and Schlangen, 2015). While these methods can achieve strong token-level detection performance, deletion-based pipelines often disrupt grammatical structure, lexical cohesion, and semantic completeness, resulting in ungrammatical or semantically broken sentences.

Disfluency detection in multilingual and low-resource contexts remain underexplored. For Indian languages, such as Hindi, Bengali, and Marathi, the existing works have focused almost exclusively on detection via Multilingual Representations for Indian Language (MuRIL)-based (Khanuja et al., 2021) sequence tagging or adversarial training (Bhat et al., 2023a; Kundu et al., 2022). However, these pipelines are limited to labeling and token deletion, offering no mechanism to *reconstruct fluent sentences*. Datasets such as DISCO (Bhat et al., 2023b) and unsupervised/semi-supervised approaches (Saini et al., 2021) have partially addressed data scarcity. However, disfluency correction in multilingual ASR settings remains unsolved. Recently, LLMs (Vaswani et al., 2017; Brown et al., 2020) are being increasingly used for coding (Jiang et al., 2026), reasoning (Huang and Chang, 2023), machine translation (Gain et al., 2026), among a wide range of other tasks. Several exploratory stud-ies have examined the role of LLMs in disfluency processing. Some of the prior works make use of LLMs as *data generators* to synthesize disfluent examples for training small detectors, while others evaluate prompt-engineered LLMs, such as GPT-4 or Gemini for direct disfluency removal. Recent studies explore disfluencies as informative cues, probing how GPT-3.5/4 handle comprehension with disfluent prompts, or leverage frameworks like Smooth-LLaMa for timestamped disfluency transcription (Altinok, 2025). However, none of these efforts integrate LLMs into a *complete disfluency correction pipeline*.

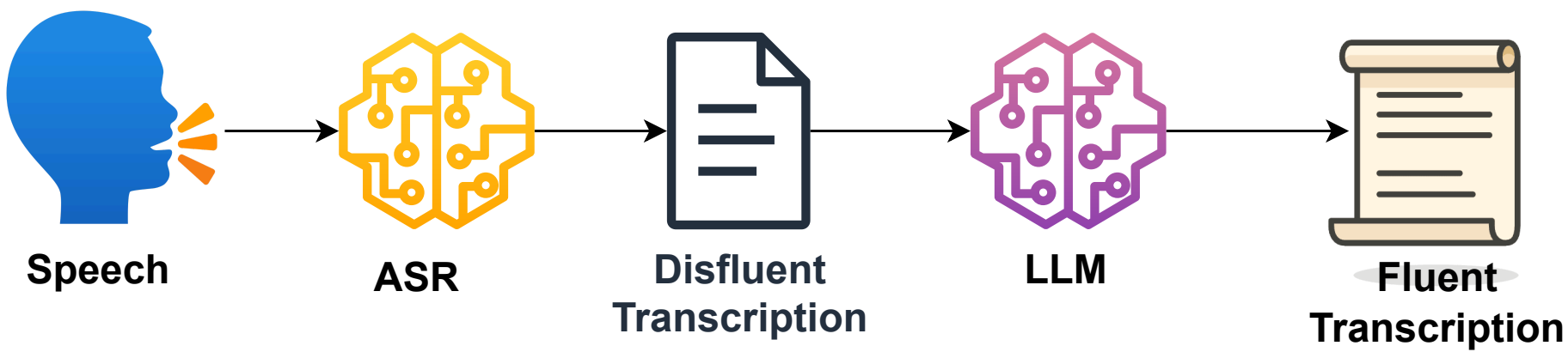

Figure 1: Disfluency correction pipeline. A potentially noisy speech transcription is first generated by an ASR model, after which a language model rewrites the disfluent text to remove disfluencies while preserving the original meaning.

Current state-of-the-art ASR systems do not resolve this problem. Our empirical analysis of Whisper v3 Large (Radford et al., 2023) and the AI4Bharat Indic Conformer (Bhogale et al., 2025) on real conversational audio reveals that approximately 30% of transcribed sentences contain at least one disfluency-including fillers, repetitions, false starts, and self-repairs-despite these systems' text normalization capabilities (Appendix A). Audio-native LLMs like Qwen2-Audio (Chu et al., 2024) have poor performance on Indic speech, producing erroneous outputs that echo in-context demonstrations verbatim. Beyond readability concerns, disfluencies cause measurable harm to downstream applications: they degrade question-answering accuracy by 0.5-1.6 points on a 5-point scale, reduce machine translation quality by 2-4.7 BLEU (Papineni et al., 2002; Post, 2018) points, and substantially degrade text-to-speech naturalness (Mean Opinion Score drops by approximately 2 points). These degradations persist even with modern instruction-tuned LLMs (Appendix B), motivating explicit disfluency correction rather than relying on implicit model robustness.

We propose a multilingual disfluency correction system that unifies detection and correction. First, we use MuRIL to tag disfluent tokens on the text obtained from ASR transcripts. Next, we use these labels to guide instruction fine-tuning of LLMs. Beyond standard cross-entropy loss, we introduce a contrastive loss that penalizes generating known disfluent tokens, encouraging fluent and faithful outputs. To our knowledge, this is the very first approach that combines detection cues, instruction tuning, and contrastive learning for multilingual disfluency correction. We evaluate on three popular Indian languages, *viz.* Hindi, Bengali, and Marathi (Kundu et al., 2022), achieving BLEU score improvement of +1.97 over non-contrastive training, +6.16 over multilingual instruction fine-tuning, and +8.54 over mBART (Liu et al., 2020). Importantly, our corrections recover the majority of the performance gap between disfluent and fluent inputs on downstream tasks like question-answering (QA), machine translation (MT) and text-to-speech synthesis (TTS) tasks (Appendix B), confirming practical utility. Our work makes the following key contributions:

- **Contrastive loss for disfluency-aware fine-tuning:** We introduce a novel multi-loss objective that combines cross-entropy with a *contrastive loss*, explicitly penalizing the model if generated outputs contain previously identified disfluent tokens.

- **First LLM-based disfluency correction pipeline:** While prior research has focused almost exclusively on detection-only models or encoder–decoder correction architectures (e.g., mBART), we are the first to design a complete *LLM-driven correction pipeline* that integrates token-level detection signals with instruction-tuned LLMs.

- **Joint detection and correction in multilingual settings:** We train MuRIL for sequence tagging to identify fluent and disfluent tokens, and crucially feed both the disfluent transcript and the predicted labels into an LLM for correction. This avoids the common pitfall of detection-only pipelines, where manual deletion of tokens can disrupt grammatical structure and semantic integrity. Our pipeline instead directly generates well-formed, fluent sentences while preserving the original meaning.

## 2 Related Work

Early research on disfluency handling primarily focused on token-level detection. End-to-end speech-to-text translation models trained directly on fluent references have been shown to implicitly drop disfluencies during decoding, though evaluation mismatches arise when fluent targets are shorter (Salesky et al., 2019). Extending this, (Jamshid Lou and Johnson, 2020) investigated whether ASR systems themselves could map disfluent audio to fluent transcripts, introducing FER/DER metrics for proper evaluation and demonstrating that integrated systems could compete with, but often lag behind, detection–deletion pipelines.

In the English setting, early work (Saini et al., 2021) framed disfluency correction as a style-transfer (translation) task, showing that semi-supervised and unsupervised approaches can remain competitive even with limited parallel data. For Indic languages, the first large-scale framework (Bhat et al., 2023a) introduced an adversarial MuRIL-based tagging model covering Hindi, Bengali, and Marathi. By leveraging synthetic and unlabeled data, it achieved notable gains in detection performance and demonstrated multilingual transfer. However, being detection-only, it relied on token deletion, often producing ungrammatical text, while the framework's rule-based synthesis failed to capture natural speech artifacts, such as false starts, truncations, and code-mixed repairs.

Zero-shot disfluency detection has also been explored for low-resource Indic languages, where MuRIL was fine-tuned on English data augmented with rule-based synthetic disfluencies (Kundu et al., 2022). This setup demonstrated robust cross-lingual transfer (F1 $>$ 70 across several disfluency types) and introduced valuable evaluation benchmarks, but the framework still operated at the token-level, relying on post-hoc deletion. As a result, performance degraded on complex speech phenomena, such as false starts or synonym repairs, and synthetic corpora diverged notably from real ASR distributions. Additional efforts like DISCO (Bhat et al., 2023b) provided well-annotated parallel data but remained limited in scale and primarily emphasized deletion or minimal editing over full-sentence correction.

With the advent of LLMs, the research focus has shifted from token-level detection to sentence-level correction and data generation. Prompt-based experiments showed that models, such as GPT-4, Claude, Gemini, and Llama, can remove disfluencies in Portuguese debates with zero- or few-shot prompting, though performance remains sensitive to prompt formulation and model size (Lima and Campelo, 2024). Other studies positioned LLMs as *data generators*, using them to create annotated disfluent sentences for training lightweight taggers that achieve state-of-the-art results without task-specific fine-tuning (Cheng et al., 2024). Insights from the Disfl-QA (Gupta et al., 2021) benchmark revealed that certain repair-based disfluencies may even aid comprehension, underscoring the need for *label-aware* rather than "delete-all" correction strategies (Rohanian et al., 2023). More recently, multimodal pipelines, such as *Smooth-LLaMa* (Altinok, 2025) have coupled a Conformer audio encoder with an LLM decoder to jointly model disfluency tokens and timestamps, pushing toward unified frameworks for detection and temporal annotation.

The proposed framework moves beyond both detection-only and prompt-only paradigms. Earlier Indic approaches have largely depended on token tagging followed by deletion (Bhat et al., 2023a; Kundu et al., 2022), often resulting in ungrammatical outputs. In contrast, our model unifies detection and correction by using MuRIL's multilingual sequence tags to guide the LLMs during instruction fine-tuning, enabling label-aware correction that preserves grammatical structure and semantic fidelity. To further address the persistence of filler repetitions and false starts, limitations typically overlooked by cross-entropy or prompt-based rewriting, we introduce a contrastive anti-disfluency loss that explicitly penalizes the regeneration of disfluent tokens.

Unlike end-to-end ASR or speech-to-text translation models that seek fluent transcripts directly from audio (Jamshid Lou and Johnson, 2020), our approach is fully ASR-agnostic and modular, facilitating integration into diverse pipelines while maintaining rigorous evaluation control. Extending disfluency correction to a multilingual Indic context (Hindi, Bengali, and Marathi), we combine detection cues, instruction tuning, and contrastive learning (Chen et al., 2020; Su et al., 2022) within a single architecture, effectively bridging token-level tagging and LLM-based generation and narrowing the gap between detection and correction.

## 3 Methodology

In this section, we describe our proposed multilingual disfluency correction framework alongside several strong baselines. We first outline the datasets and pre-processing steps used, followed by baseline models such as mBART and instruction-tuned LLMs. Finally, we describe in detail our method, which integrates MuRIL-based token tagging with instruction fine-tuning and a contrastive loss for fluent correction.

### 3.1 Dataset

We use the parallel fluent-disfluent dataset (Kundu et al., 2022) for Hindi, Bengali, and Marathi. This is, to our knowledge, the only publicly available resource that provides parallel data for disfluency detection and correction in these three Indian languages. The corpus extends PMIndia - A Collection of Parallel Corpora for the Languages of India (Haddow and Kirefu, 2020) by algorithmically inducing fillers, repetitions, corrections, and false starts into fluent text, followed by manual refinement by native speakers to ensure naturalness. Additional manually transcribed disfluent data provide a more realistic evaluation set. In total, the dataset comprises roughly 120k parallel sentence pairs (40k per language), supporting both tagging and supervised correction tasks. The detailed statistics are provided in Table 1.

| Language | Synthetic Pairs | Manually Edited | Real Test |
|---|---|---|---|
| Hindi | ∼40,000 | 575 sentences | 150 sentences |
| Bengali | ∼40,000 | 500 sentences | 300 sentences |
| Marathi | ∼40,000 | 420 sentences | 250 sentences |
| **Total** | ∼120,000 | 1,495 sentences | 700 sentences |

Table 1: Statistics of the parallel disfluent–fluent datasets used for training and evaluation. Synthetic data is generated via rule-based induction; manually edited and real datasets serve as evaluation benchmarks.

**Train–validation–test splits:** For multilingual fine-tuning, we combine the data from all three languages and split it into 80% training, 10% validation, and 10% test sets. For monolingual fine-tuning, we prepare separate splits per language following the same 80-10-10 strategy.

### 3.2 Pre-processing

For MuRIL-based tagging, token-level fluent/disfluent labels are derived by aligning parallel fluent–disfluent pairs and marking tokens absent in the fluent version as disfluent, generating supervised labels for sequence tagging. For LLM fine-tuning, the data are formatted in an Alpaca-style instruction–input–output schema (Taori et al., 2023), where the instruction defines the disfluency removal task, the input provides the disfluent sentence (optionally augmented with MuRIL labels), and the output is the corresponding fluent text. In the contrastive variant, the set of disfluent tokens identified through alignment is additionally supplied as auxiliary input, and the model is optimized with a combined objective: cross-entropy loss for fluency generation and a contrastive penalty discouraging regeneration of known disfluent tokens.

We evaluate on two test sets per language:
(1) Manually edited synthetic data, which is high-quality but synthetic in origin.
(2) Real disfluent transcripts, transcribed from conversational interviews in Hindi, Bengali, and Marathi.
These are the only available parallel resources for disfluency correction in these languages, making them a critical benchmark for our study.

### 3.3 Baselines

We compare our method against several strong baselines spanning supervised encoder–decoder models and instruction-tuned LLMs. These serve to benchmark the effectiveness of multilingual fine-tuning and contrastive learning for disfluency correction.

#### 3.3.1 Fine-tuned mBART

Prior works on English-only studies have benchmarked encoder–decoder models (Seq2Seq, BART/mBART) against unsupervised and semi-supervised methods (Saini et al., 2021), framing disfluency correction as a translation task from disfluent to fluent text. However, evaluation has largely been restricted to English corpora such as Switchboard.

In our work, we establish a multilingual mBART baseline for Hindi, Bengali, and Marathi. We fine-tune mBART (Liu et al., 2020) using the parallel disfluent–fluent pairs (Kundu et al., 2022), combining all three languages into a single multilingual training setup. We evaluate the finetuned mBART checkpoints on the manually edited and real test sets for each language. The results reported in Table 3 serve as a strong supervised encoder–decoder baseline.

### 3.3.2 Zero-Shot Prompting with LLMs

We evaluate a zero-shot prompting baseline using the Llama-3.2-3B-Instruct (Grattafiori et al., 2024) and Qwen2.5-3B-Instruct (Team, 2024) model. In this setting, the model is directly prompted to rewrite disfluent sentences into fluent ones without any task-specific fine-tuning or access to token-level tags. This setup assesses the inherent ability of instruction-tuned LLMs to perform disfluency correction purely through prompting.

### 3.3.3 Multilingual Instruction Tuning

We evaluate the effectiveness of LLMs in correcting disfluencies when fine-tuned with disfluency-specific instructions but *without access to token-level disfluency signals*. We frame disfluency correction as a supervised instruction-tuning task, where the model is given a natural-language prompt together with a disfluent input sentence, and asked to generate its fluent counterpart. The model is then fine-tuned with disfluent–fluent pairs, where the input is the concatenation of the instruction and a disfluent sentence, and the output is the corresponding fluent reference. Unlike detection-driven pipelines, this formulation requires the model to implicitly identify and remove disfluent spans while generating fluent sentences.

## 3.4 Proposed Method: Disfluency Correction with Tags

The first stage of our pipeline involves detecting disfluent tokens in ASR transcripts using a multilingual encoder-based sequence tagging model. The goal of this stage is not standalone detection accuracy, but rather to provide disfluency-aware token-level signals that can guide downstream LLM-based correction. We adopt MuRIL (Khanuja et al., 2021), a transformer-based encoder pretrained on 17 Indian languages, as the backbone for token classification. The model is fine-tuned in a sequence tagging setting, where each token is classified as either fluent or disfluent. The label set is defined as: (0 $\rightarrow$ fluent, 1 $\rightarrow$ disfluent). Each input sentence is first tokenized using MuRIL's WordPiece tokenizer. The disfluency annotations are aligned at the token-level such that each subword token inherits the label of its parent word. Training data consists of disfluent ASR transcripts paired with token-level annotations indicating whether each word is fluent or disfluent. This setup allows the model to learn

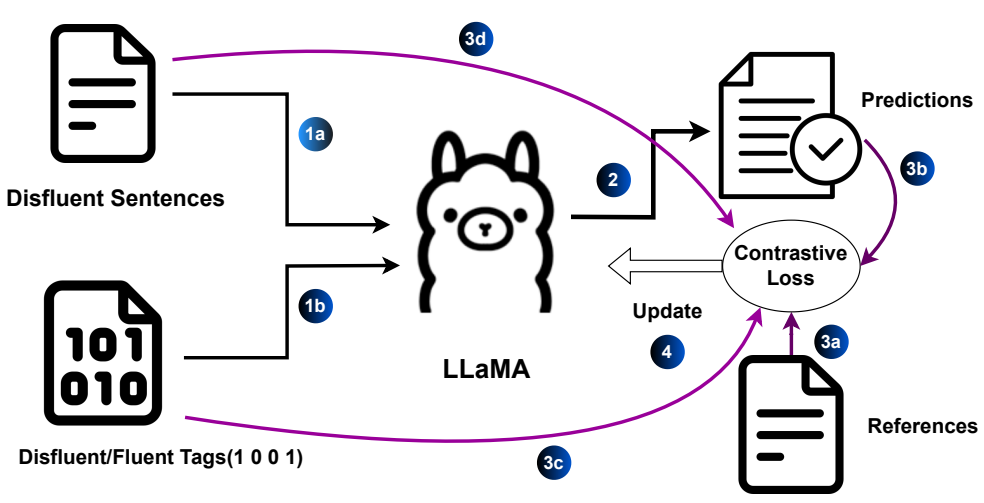

Figure 2: Flow diagram of the proposed multilingual disfluency correction pipeline integrating MuRIL-based token tagging with LLM fine-tuning. The model jointly optimizes cross-entropy and contrastive losses to suppress disfluent tokens and generate fluent text.

context-sensitive disfluency cues, including repetitions, fillers, and false starts.

**Training Setup:** We fine-tune MuRIL jointly on Hindi, Marathi, and Bengali in a multilingual setting, enabling the model to share disfluency representations across diverse languages. The training objective is the standard cross-entropy loss over token labels:

$$\mathcal{L}_{\text{detect}} = -\sum_{i=1}^{N}\sum_{t=1}^{T_i} \log P_\theta(y_i^t \mid x_i), \qquad (1)$$

where $x_i$ is the input sentence, $T_i$ is its sentence length, $y_i^t \in \{0,1\}$ is the gold label for token $t$, and $P_\theta$ is the MuRIL classifier's output.

Our methodology integrates token-level disfluency signals predicted by a MuRIL with LLM based correction. This design moves beyond token-level deletion pipelines and explicitly conditions generation on detected labels. For each sentence, we take
(i) the ASR-generated disfluent transcript,
(ii) token-level labels predicted by MuRIL (fluent or disfluent), and
(iii) the corresponding fluent reference. During training, we concatenate the natural-language instruction, the disfluent input sentence, and the MuRIL-provided token and label sequence. The target output is the fluent reference sentence. This setup requires the LLM not only to drop disfluent tokens but also to generate the fluent sentence.

**Objective Function:** We optimize the model using the cross-entropy loss between the generated fluent output and the gold fluent reference. Given a training dataset $\mathcal{D} = \{(x_i, y_i)\}_{i=1}^{N}$, where $x_i$ is the input (instruction + disfluent sentence + MuRIL signals) and $y_i = (y_i^1, \ldots, y_i^T)$ is the gold fluent sentence, the loss is:

$$\mathcal{L}_{\text{CE}} = -\sum_{i=1}^{N} \sum_{t=1}^{T} \log P_\theta(y_i^t \mid y_i^{<t}, x_i). \quad (2)$$

Here:

- $P_\theta$ denotes the conditional probability distribution defined by the LLM model with parameters $\theta$,
- $y_i^t$ is the $t$-th token in the gold fluent sentence,
- $y_i^{<t}$ denotes the gold prefix $(y_i^1, \ldots, y_i^{t-1})$,
- $x_i$ is the combined input including instruction, disfluent sentence, and MuRIL-provided disfluency labels.

| Dataset | Precision | Recall | F1-score | Sentence Acc. |
|---|---|---|---|---|
| **Manually Edited Data** | | | | |
| Bengali | 0.9774 | 0.9977 | 0.9874 | 0.8580 |
| Hindi | 0.9785 | 0.9977 | 0.9880 | 0.8383 |
| Marathi | 0.9776 | 0.9954 | 0.9864 | 0.8548 |
| **Real Data** | | | | |
| Bengali | 0.8642 | 0.9930 | 0.9241 | 0.3267 |
| Hindi | 0.9565 | 0.9837 | 0.9699 | 0.6333 |
| Marathi | 0.9068 | 0.9524 | 0.9290 | 0.5040 |

Table 2: Comparison of disfluency detection performance of MuRIL-based sequence tagger on manually edited and real ASR data across three Indian languages (Hindi, Bengali, and Marathi).

This formulation ensures the model learns to generate the exact fluent reference sequence, thereby implicitly removing disfluencies while preserving the grammatical and semantic content.

Table 2 summarizes the performance of the MuRIL-based disfluency detection model. While the detector achieves strong token-level precision and recall on manually edited data, it remains competitive on real ASR outputs, particularly at the sentence-level, despite increased transcription noise. Importantly, our goal is not to maximize standalone detection accuracy, but to leverage these token-level disfluency signals as auxiliary supervision for downstream LLM-based correction.

### 3.5 Proposed Method: Disfluency Correction with Contrastive Loss

While subsection 3.4 demonstrated the effectiveness of instruction-tuned correction with detection signals, we observe that models trained solely with cross-entropy can occasionally reproduce disfluent words present in the input, despite receiving explicit token-level labels. To address this limitation, we introduce a novel contrastive correction objective that directly penalizes the generation of disfluent tokens.

**Contrastive Objective:** We design a multi-loss setup that combines the standard cross-entropy objective with a contrastive penalty. The intuition is straightforward: if any generated output token matches one of the known disfluent tokens, the model incurs an additional penalty. This explicitly forces the model to avoid reproducing disfluencies, complementing the implicit learning signal provided by cross-entropy. Formally, the total loss is defined as:

$$\mathcal{L}_{\text{contrastive}} = \frac{1}{N} \sum_{i=1}^{N} \frac{1}{T_i} \sum_{t=r_i}^{T_i} \left[ -\log \left( 1 - s_{i,t} \right) \right], \quad (3)$$

$$\text{where } s_{i,t} = \sum_{v \in D_i} w_v \, P_\theta(v \mid y_i^{<t}, x_i). \quad (4)$$

Here:

- $N$ is the number of training examples in a batch and
- $T_i$ denotes the total number of tokens in the $i$-th target sequence.
- $r_i$ is the index from which the response tokens begin (i.e., positions after the instruction and input).
- $D_i$ represents the set of disfluent tokens identified for the $i$-th example.
- $w_v \in (0, 1]$ is the geometric decay weight assigned to sub-tokens of a disfluent word (e.g., $1, 0.5, 0.25, \ldots$).
- $P_\theta(v \mid y_i^{<t}, x_i)$ denotes the model's predicted probability of token $v$ at timestep $t$, given the prefix $y_i^{<t}$ and input $x_i$.
- $x_i$ is the combined input including the instruction, disfluent sentence, and MuRIL-provided disfluency labels.

$$\mathcal{L}_{\text{total}} = \mathcal{L}_{\text{CE}} + \lambda \cdot \mathcal{L}_{\text{contrastive}}, \quad (5)$$

where $\mathcal{L}_{\text{CE}}$ is the cross-entropy loss between the generated sequence and the fluent reference,

| Language | Dataset | Multilingual Instruction F/t | | | mBART | | | W/o Contrastive Loss | | | With Contrastive Loss | | |
|---|---|---|---|---|---|---|---|---|---|---|---|---|---|
| | | BLEU | chrF2 | TER | BLEU | chrF2 | TER | BLEU | chrF2 | TER | BLEU | chrF2 | TER |
| Hindi | Manually Edited | 92.5 | 98.0 | 6.1 | 85.7 | 91.3 | 8.3 | 93.8 | 97.6 | 4.8 | **94.6** | **98.5** | **4.0** |
| | Real Data | 64.8 | 81.7 | 23.4 | 71.4 | 85.5 | 15.1 | 87.4 | 93.3 | 9.2 | **90.4** | **95.6** | **5.8** |
| Bengali | Manually Edited | 94.2 | 98.1 | 4.7 | 84.3 | 93.0 | 8.5 | 92.6 | 98.0 | 5.7 | **94.8** | **98.8** | **4.1** |
| | Real Data | 69.6 | 89.0 | 21.6 | 73.5 | 87.9 | **13.0** | 70.7 | 90.5 | 20.8 | **74.4** | **93.8** | 17.9 |
| Marathi | Manually Edited | 94.4 | 98.4 | 4.5 | 83.7 | 92.1 | 9.2 | 93.0 | 98.0 | 5.1 | **94.7** | **98.8** | **4.0** |
| | Real Data | 80.0 | 94.3 | 11.8 | 82.6 | 93.1 | **8.2** | 83.2 | 95.5 | 9.3 | **83.6** | **96.6** | 9.2 |

Table 3: Performance comparison of multilingual disfluency correction strategies on Hindi, Bengali, and Marathi using the Llama-3.2-3B-Instruct model. We compare multilingual instruction fine-tuning, mBART, training without contrastive loss, and the proposed contrastive approach on manually edited and real ASR data. Scores are reported using BLEU, $\text{chrF}_2$ (↑), and TER (↓) against gold fluent references.

| Language | Dataset | Multilingual Instruction F/t | | | W/o Contrastive Loss | | | With Contrastive Loss | | |
|---|---|---|---|---|---|---|---|---|---|---|
| | | BLEU | chrF2 | TER | BLEU | chrF2 | TER | BLEU | chrF2 | TER |
| Hindi | Manually Edited | 92.3 | 94.3 | 9.8 | 93.0 | 96.5 | 6.1 | **96.1** | **98.0** | **3.7** |
| | Real Data | 79.5 | 92.9 | 13.8 | 84.2 | 92.7 | 9.5 | **91.1** | **94.9** | **6.0** |
| Bengali | Manually Edited | 88.2 | 96.5 | 11.7 | 90.1 | 97.1 | 6.7 | **96.4** | **98.6** | **3.4** |
| | Real Data | 66.2 | 87.3 | 29.4 | 71.4 | 88.3 | 20.3 | **75.9** | **93.2** | **16.2** |
| Marathi | Manually Edited | 93.3 | 98.4 | 4.8 | 92.0 | 97.2 | 5.8 | **95.1** | **98.4** | **3.8** |
| | Real Data | 78.7 | 88.2 | 17.7 | 80.2 | 91.4 | 14.0 | **84.4** | **94.3** | **10.0** |

Table 4: Performance comparison of multilingual disfluency correction strategies on Hindi, Bengali, and Marathi using the Qwen2.5-3B-Instruct model. Results are reported on manually edited and real ASR data using BLEU, $\text{chrF}_2$ (↑), and TER (↓).

$\mathcal{L}_{\text{contrastive}}$ penalizes the generation probability of disfluent tokens, and $\lambda$ is a weight with warm-up scheduling.

Figure 2 illustrates the proposed multilingual disfluency correction training pipeline. In Step 1a, disfluent sentences are fed into LLM along with their corresponding token-level fluent/disfluent tags obtained from MuRIL in Step 1b. These two signals jointly condition the model to identify and suppress disfluent regions during generation. The model produces a fluent hypothesis (Step 2 $\rightarrow$ 3b), which is compared against the ground-truth fluent reference (Step 3a) to compute the *cross-entropy loss*. Simultaneously, the original disfluent input (Step 3d) and its MuRIL-derived disfluency tags (Step 3c) are used to extract known disfluent tokens, which are then penalized under a *contrastive loss* that discourages their regeneration. Intuitively, cross-entropy provides a positive signal that aligns the model with fluent targets but does not explicitly teach it to avoid regenerating disfluent spans. The contrastive term supplies the missing negative signal by pushing representations of MuRIL-tagged disfluencies away from the fluent target space, reducing their likelihood during decoding. Together, this push–pull dynamic yields a generator that not only imitates fluent structure and meaning but also actively resists common disfluency patterns across languages. The combined objective-jointly optimizing cross-entropy for fluency alignment and contrastive loss for disfluency suppression-is backpropagated in Step 4 to update weights of the LLM, thereby reinforcing fluent, structure-preserving correction.

## 4 Results and Analysis

Table 3 and Table 4 report results for four supervised strategies: *mBART*, *Multilingual Instruction Fine-tuning*, *Without Contrastive Loss*, and the *Proposed Contrastive Loss* approach. Experiments are conducted on Hindi, Bengali, and Marathi using both manually edited and real ASR transcripts, evaluated with two instruction-tuned LLM backbones: Llama-3.2-3B-Instruct and Qwen2.5-3B-Instruct. We further analyze multilingual versus language-specific instruction fine-tuning and zero-shot cross-lingual transfer across languages in Appendix D.

The mBART baseline performs competitively, and for the Llama-3.2-3B backbone, on real ASR inputs it can achieve higher BLEU than multilingual instruction fine-tuning (Table 3), likely reflecting its conservative, denoising-style seq2seq behavior under noisy inputs. In contrast, for the Qwen-2.5-3B backbone, multilingual instruction fine-tuning already surpasses mBART, indicating stronger ro-bustness of instruction-tuned LLMs with better multilingual grounding. However, instruction fine-tuning provides a flexible correction framework, and conditioning the LLM on token-level disfluency cues via MuRIL tagging (without contrastive loss) consistently improves over multilingual instruction fine-tuning by guiding rewriting toward detected disfluent spans. Notably, even this non-contrastive, MuRIL-tag-conditioned variant is competitive relative to mBART, and in many settings the MuRIL-conditioned LLaMA/Qwen models outperform mBART, indicating that explicit disfluency-aware supervision helps unlock the stronger contextual modeling capacity of instruction-tuned LLMs for correction. Importantly, adding the proposed contrastive objective yields the best overall performance across languages and datasets, substantially surpassing both mBART and the non-contrastive variants.

| LLM-as-a-Judge: Proposed Approach vs. Parallel Fine-tuning | | | | |
|---|---|---|---|---|
| **Language** | **Dataset** | **Proposed (%)** | **Parallel (%)** | **Draw (%)** |
| Hindi | Manually Edited | 18.1 | 7.3 | 74.6 |
| | Real Data | 28.0 | 9.3 | 62.7 |
| Marathi | Manually Edited | 14.0 | 7.6 | 78.3 |
| | Real Data | 30.0 | 8.0 | 62.0 |
| Bengali | Manually Edited | 17.4 | 27.2 | 55.4 |
| | Real Data | 18.0 | 27.0 | 55.0 |

Table 5: Percentage-wise evaluation of the Proposed Approach vs. Parallel Fine-tuning using LLM-as-a-Judge.

The proposed contrastive variant achieves the best overall performance across both architectures, showing consistent improvements across metrics and languages. When compared against the non-contrastive baseline and averaged over all languages and evaluation settings, contrastive training improves Llama-3.2-3B-Instruct by +1.97 BLEU and +1.53 chrF2 (Popović, 2015), while reducing TER (Snover et al., 2006) by 1.65 points. The gains are even more pronounced for Qwen2.5-3B-Instruct, where contrastive supervision yields average improvements of +4.68 BLEU and +2.37 chrF2, along with a 3.22-point reduction in TER.

Comparing the two architectures, Qwen2.5-3B-Instruct benefits more strongly from the proposed contrastive objective, achieving an additional +2.71 BLEU and +0.84 chrF2 improvement, and a further 1.57-point TER reduction over Llama-3.2-3B-Instruct on average. This indicates that the proposed contrastive learning framework generalizes across model families, while being particularly effective for instruction-tuned models with stronger generative priors.

For example, on the Hindi manually edited set using Qwen2.5-3B-Instruct, the contrastive model achieves 96.1 BLEU, 98.0 chrF2, and 3.7 TER, outperforming all competing systems. Even on challenging real ASR data, it remains robust (91.1 BLEU / 94.9 chrF2 / 6.0 TER), demonstrating that contrastive supervision effectively suppresses fillers, repetitions, and false starts while preserving grammatical and semantic fidelity.

To assess the output quality, we additionally perform evaluation using the LLM-as-a-Judge (Zheng et al., 2023; Li et al., 2025) framework (Table 5), employing `Qwen2.5-3B-Instruct` as the evaluator. We select Qwen instead of LLaMA to mitigate potential self-preference biases (Wataoka et al., 2025), which could otherwise inflate the evaluation scores. Moreover, prior work has highlighted that LLM judges can exhibit positional biases (Bito et al., 2025), favoring candidate outputs based on their order of presentation. To reduce such bias, we conduct pairwise comparisons in both directions and report the averaged results.

The judge model compares outputs from our proposed system and the parallel fine-tuned baseline, rating each on fluency and meaning preservation. The contrastive model is preferred in most cases, particularly for real conversational data (e.g., 28% vs. 9% for Hindi and 30% vs. 8% for Marathi). These findings align with the quantitative metrics, confirming that contrastive loss produces more natural, coherent, and human-like corrections.

Other baselines, such as zero-shot disfluency correction with pretrained LLMs perform poorly and we report the detailed results in Appendix E. Overall, the integration of contrastive loss into multilingual instruction tuning achieves significant improvements across both automatic and human evaluation, establishing it as a robust framework for fluency correction in multilingual speech-derived text.

We further compare against frontier proprietary models, GPT-4o (OpenAI et al., 2024) and Gemini 2.5 Pro (Comanici et al., 2025) in Appendix I. GPT-4o is evaluated under zero-shot, few-shot, and zero-shot CoT prompting strategies, while Gemini 2.5 Pro is evaluated under zero-shot prompting on the real test set only. Our 3B contrastive model matches or outperforms GPT-4o on the majority of evaluation conditions and substantially outperforms Gemini 2.5 Pro across all three languages.

We further conducted a human evaluation to assess fluency and meaning preservation of the corrected outputs. As summarized in Appendix H, native-language experts consistently preferred outputs generated using the proposed contrastive approach over the multilingual instruction-finetuned Llama-3.2-3B-Instruct baselines across all three languages, corroborating trends observed in the automatic and LLM-as-a-Judge evaluations.

## 5 Conclusion

In this paper, we have introduced a multilingual disfluency correction framework that combines token-level detection signals with instruction-tuned LLMs and a contrastive learning objective. Experiments on three popular Indian languages, namely Hindi, Bengali, and Marathi show that the proposed approach consistently surpasses strong baselines such as mBART and multilingual fine-tuning, achieving notable gains in BLEU and chrF2 while reducing TER. The contrastive objective effectively penalizes the regeneration of fillers and repetitions, leading to fluent and meaning-preserving outputs. LLM-as-a-Judge evaluation further validates these improvements, with the proposed model preferred for its naturalness and coherence. Importantly, our downstream task evaluation (Appendix B) demonstrates that correction quality translates directly to application performance: our corrected outputs recover the majority of the performance gap between disfluent and gold-fluent inputs across QA, MT, and TTS tasks, confirming practical utility in real-world deployments. Notably, our approach matches or surpasses GPT-4o on 4 out of 6 evaluation conditions and substantially outperforms Gemini 2.5 Pro across all three languages (Appendix I), confirming that task-specific contrastive training cannot be substituted by scaling or advanced prompting alone. Overall, contrastive supervision emerges as a simple yet powerful enhancement to instruction tuning, offering a scalable and language-agnostic solution for fluent text generation in multilingual and speech-driven NLP systems.

## Limitations

Our work has a few limitations and caveats. We evaluate two open-source model families, Llama-3.2-3B-Instruct and Qwen2.5-3B-Instruct, both at the 3B-parameter scale, due to computational constraints; larger models could potentially yield stronger results, but exploring such scaling effects lies beyond this study. Nevertheless, our methodology is model-agnostic and can, in principle, extend to larger or closed-source LLMs. While closed-source evaluators might offer more accurate LLM-as-judge performance, we deliberately employ open-source evaluation models to ensure reproducibility and transparency, acknowledging a possible trade-off in judgment fidelity. Our work also faces data-related constraints: only one publicly available parallel disfluent–fluent dataset currently exists for Hindi, Bengali, and Marathi, limiting both language and domain coverage. Real test sets remain small, as manual disfluency annotation is costly, constraining the statistical robustness of per-language analyses. Moreover, prior research in Indic NLP focuses primarily on *disfluency detection* (Kundu et al., 2022), with no available baselines for full-span *disfluency correction*. Consequently, comparative evaluation remains limited. Despite these constraints, our contrastive instruction-tuning framework achieves consistent fluency improvements across multiple languages, underscoring its robustness and adaptability in truly low-resource, data-scarce settings.

## Acknowledgement

Baban Gain gratefully acknowledge the Prime Minister's Research Fellowship (PMRF) program for providing financial support and enabling this research. The authors gratefully acknowledge Dr. Saroj Kumar Jha, Anansa Roy, Dr. Saloka Sengupta, Souravi Halder, Ayodhya Murmu and Dr. Prafull Meshram for carrying out the manual evaluation. The authors gratefully acknowledge the project "Centre of Indian Language Data (COIL-D)" under the flagship mission of Bhashini, funded by MeitY, Government of India, for the financial grant that enabled the successful conduct of this research.

## References

Duygu Altinok. 2025. Smooth operators: Llms translating imperfect hints into disfluency-rich transcripts. *Preprint*, arXiv:2506.18510.

Vineet Bhat, Preethi Jyothi, and Pushpak Bhattacharyya. 2023a. Adversarial training for low-resource disfluency correction. In *Findings of the Association for Computational Linguistics: ACL 2023*, pages 8112–8122, Toronto, Canada. Association for Computational Linguistics.

Vineet Bhat, Preethi Jyothi, and Pushpak Bhattacharyya. 2023b. DISCO: A large scale human annotated corpus for disfluency correction in Indo-European languages. In *Findings of the Association for Computational Linguistics: EMNLP 2023*, pages 12833–12857, Singapore. Association for Computational Linguistics.

Kaushal Santosh Bhogale, Deovrat Mehendale, Tahir Javed, Devbrat Anuragi, Sakshi Joshi, Sai Sundaresan, Aparna Ananthanarayanan, Sharmistha Dey, Sathish Kumar Reddy G, Anusha Srinivasan, Abhigyan Raman, Pratyush Kumar, and Mitesh M. Khapra. 2025. Towards bringing parity in pretraining datasets for low-resource indian languages. In *ICASSP 2025 - 2025 IEEE International Conference on Acoustics, Speech and Signal Processing (ICASSP)*, pages 1–5.

Ethan Bito, Yongli Ren, and Estrid He. 2025. Evaluating position bias in large language model recommendations. *Preprint*, arXiv:2508.02020.

Tom B. Brown, Benjamin Mann, Nick Ryder, Melanie Subbiah, Jared Kaplan, Prafulla Dhariwal, Arvind Neelakantan, Pranav Shyam, Girish Sastry, Amanda Askell, Sandhini Agarwal, Ariel Herbert-Voss, Gretchen Krueger, Tom Henighan, Rewon Child, Aditya Ramesh, Daniel M. Ziegler, Jeffrey Wu, Clemens Winter, and 12 others. 2020. Language models are few-shot learners. *Preprint*, arXiv:2005.14165.

Ting Chen, Simon Kornblith, Mohammad Norouzi, and Geoffrey Hinton. 2020. A simple framework for contrastive learning of visual representations. In *Proceedings of the 37th International Conference on Machine Learning*, ICML'20. JMLR.org.

Zhenrong Cheng, Jiayan Guo, Hao Sun, and Yan Zhang. 2024. Boosting disfluency detection with large language model as disfluency generator. *Preprint*, arXiv:2403.08229.

Yunfei Chu, Jin Xu, Qian Yang, Haojie Wei, Xipin Wei, Zhifang Guo, Yichong Leng, Yuanjun Lv, Jinzheng He, Junyang Lin, Chang Zhou, and Jingren Zhou. 2024. Qwen2-audio technical report. *Preprint*, arXiv:2407.10759.

Gheorghe Comanici, Eric Bieber, Mike Schaekermann, Ice Pasupat, Noveen Sachdeva, Inderjit Dhillon, Marcel Blistein, Ori Ram, Dan Zhang, Evan Rosen, Luke Marris, Sam Petulla, Colin Gaffney, Asaf Aharoni, Nathan Lintz, Tiago Cardal Pais, Henrik Jacobsson, Idan Szpektor, Nan-Jiang Jiang, and 3416 others. 2025. Gemini 2.5: Pushing the frontier with advanced reasoning, multimodality, long context, and next generation agentic capabilities. *Preprint*, arXiv:2507.06261.

Baban Gain, Dibyanayan Bandyopadhyay, Asif Ekbal, and Trilok Nath Singh. 2026. Bridging the linguistic divide: A survey on leveraging large language models for machine translation. *Preprint*, arXiv:2504.01919.

Aaron Grattafiori, Abhimanyu Dubey, Abhinav Jauhri, Abhinav Pandey, Abhishek Kadian, Ahmad Al-Dahle, Aiesha Letman, Akhil Mathur, Alan Schelten, Alex Vaughan, Amy Yang, Angela Fan, Anirudh Goyal, Anthony Hartshorn, Aobo Yang, Archi Mitra, Archie Sravankumar, Artem Korenev, Arthur Hinsvark, and 542 others. 2024. The llama 3 herd of models. *Preprint*, arXiv:2407.21783.

Aditya Gupta, Jiacheng Xu, Shyam Upadhyay, Diyi Yang, and Manaal Faruqui. 2021. Disfl-qa: A benchmark dataset for understanding disfluencies in question answering. *Preprint*, arXiv:2106.04016.

Barry Haddow and Faheem Kirefu. 2020. Pmindia – a collection of parallel corpora of languages of india. *Preprint*, arXiv:2001.09907.

Matthew Honnibal and Mark Johnson. 2014. Joint incremental disfluency detection and dependency parsing. *Transactions of the Association for Computational Linguistics*, 2:131–142.

Julian Hough and David Schlangen. 2015. Recurrent neural networks for incremental disfluency detection. In *Interspeech 2015*, pages 849–853.

Jie Huang and Kevin Chen-Chuan Chang. 2023. Towards reasoning in large language models: A survey. In *Findings of the Association for Computational Linguistics: ACL 2023*, pages 1049–1065, Toronto, Canada. Association for Computational Linguistics.

Paria Jamshid Lou and Mark Johnson. 2020. End-to-end speech recognition and disfluency removal. In *Findings of the Association for Computational Linguistics: EMNLP 2020*, pages 2051–2061, Online. Association for Computational Linguistics.

Juyong Jiang, Fan Wang, Jiasi Shen, Sungju Kim, and Sunghun Kim. 2026. A survey on large language models for code generation. *ACM Trans. Softw. Eng. Methodol.*, 35(2).

Mark Johnson and Eugene Charniak. 2004. A TAG-based noisy-channel model of speech repairs. In *Proceedings of the 42nd Annual Meeting of the Association for Computational Linguistics (ACL-04)*, pages 33–39, Barcelona, Spain.

Simran Khanuja, Diksha Bansal, Sarvesh Mehtani, Savya Khosla, Atreyee Dey, Balaji Gopalan, Dilip Kumar Margam, Pooja Aggarwal, Rajiv Teja Nagipogu, Shachi Dave, Shruti Gupta, Subhash Chandra Bose

Gali, Vish Subramanian, and Partha Talukdar. 2021. Muril: Multilingual representations for indian languages. *Preprint*, arXiv:2103.10730.

Rohit Kundu, Preethi Jyothi, and Pushpak Bhattacharyya. 2022. Zero-shot disfluency detection for Indian languages. In *Proceedings of the 29th International Conference on Computational Linguistics*, pages 4442–4454, Gyeongju, Republic of Korea. International Committee on Computational Linguistics.

Yoach Lacombe, Vaibhav Srivastav, and Sanchit Gandhi. 2024. Parler-tts. `https://github.com/huggingface/parler-tts`.

Dawei Li, Bohan Jiang, Liangjie Huang, Alimohammad Beigi, Chengshuai Zhao, Zhen Tan, Amrita Bhattacharjee, Yuxuan Jiang, Canyu Chen, Tianhao Wu, Kai Shu, Lu Cheng, and Huan Liu. 2025. From generation to judgment: Opportunities and challenges of LLM-as-a-judge. In *Proceedings of the 2025 Conference on Empirical Methods in Natural Language Processing*, pages 2757–2791, Suzhou, China. Association for Computational Linguistics.

Pedro Siqueira de Lima and Claudio Campelo. 2024. Disfluency detection and removal in speech transcriptions via large language models. In *Proceedings of the 15th Brazilian Symposium in Information and Human Language Technology*, pages 156–164, Belém do Pará, Brazil. Association for Computational Linguistics.

Yinhan Liu, Jiatao Gu, Naman Goyal, Xian Li, Sergey Edunov, Marjan Ghazvininejad, Mike Lewis, and Luke Zettlemoyer. 2020. Multilingual denoising pre-training for neural machine translation. *Transactions of the Association for Computational Linguistics*, 8:726–742.

Dan Lyth and Simon King. 2024. Natural language guidance of high-fidelity text-to-speech with synthetic annotations. *Preprint*, arXiv:2402.01912.

OpenAI, Josh Achiam, Steven Adler, Sandhini Agarwal, Lama Ahmad, Ilge Akkaya, Florencia Leoni Aleman, Diogo Almeida, Janko Altenschmidt, Sam Altman, Shyamal Anadkat, Red Avila, Igor Babuschkin, Suchir Balaji, Valerie Balcom, Paul Baltescu, Haiming Bao, Mohammad Bavarian, Jeff Belgum, and 262 others. 2024. Gpt-4 technical report. *Preprint*, arXiv:2303.08774.

Kishore Papineni, Salim Roukos, Todd Ward, and Wei-Jing Zhu. 2002. Bleu: a method for automatic evaluation of machine translation. In *Proceedings of the 40th Annual Meeting of the Association for Computational Linguistics*, pages 311–318, Philadelphia, Pennsylvania, USA. Association for Computational Linguistics.

Maja Popović. 2015. chrF: character n-gram F-score for automatic MT evaluation. In *Proceedings of the Tenth Workshop on Statistical Machine Translation*, pages 392–395, Lisbon, Portugal. Association for Computational Linguistics.

Matt Post. 2018. A call for clarity in reporting BLEU scores. In *Proceedings of the Third Conference on Machine Translation: Research Papers*, pages 186–191, Brussels, Belgium. Association for Computational Linguistics.

Vineel Pratap, Andros Tjandra, Bowen Shi, Paden Tomasello, Arun Babu, Sayani Kundu, Ali Elkahky, Zhaoheng Ni, Apoorv Vyas, Maryam Fazel-Zarandi, Alexei Baevski, Yossi Adi, Xiaohui Zhang, Wei-Ning Hsu, Alexis Conneau, and Michael Auli. 2023. Scaling speech technology to 1,000+ languages. *Preprint*, arXiv:2305.13516.

Alec Radford, Jong Wook Kim, Tao Xu, Greg Brockman, Christine McLeavey, and Ilya Sutskever. 2023. Robust speech recognition via large-scale weak supervision. In *Proceedings of the 40th International Conference on Machine Learning*, ICML'23. JMLR.org.

Mohammad Sadegh Rasooli and Joel Tetreault. 2013. Joint parsing and disfluency detection in linear time. In *Proceedings of the 2013 Conference on Empirical Methods in Natural Language Processing*, pages 124–129, Seattle, Washington, USA. Association for Computational Linguistics.

Morteza Rohanian, Farhad Nooralahzadeh, Omid Rohanian, David Clifton, and Michael Krauthammer. 2023. Disfluent cues for enhanced speech understanding in large language models. In *Findings of the Association for Computational Linguistics: EMNLP 2023*, pages 3676–3684, Singapore. Association for Computational Linguistics.

Nikhil Saini, Drumil Trivedi, Shreya Khare, Tejas Dhamecha, Preethi Jyothi, Samarth Bharadwaj, and Pushpak Bhattacharyya. 2021. Disfluency correction using unsupervised and semi-supervised learning. In *Proceedings of the 16th Conference of the European Chapter of the Association for Computational Linguistics: Main Volume*, pages 3421–3427, Online. Association for Computational Linguistics.

Elizabeth Salesky, Matthias Sperber, and Alexander Waibel. 2019. Fluent translations from disfluent speech in end-to-end speech translation. In *Proceedings of the 2019 Conference of the North American Chapter of the Association for Computational Linguistics: Human Language Technologies, Volume 1 (Long and Short Papers)*, pages 2786–2792, Minneapolis, Minnesota. Association for Computational Linguistics.

Ashwin Sankar, Yoach Lacombe, Sherry Thomas, Praveen Srinivasa Varadhan, Sanchit Gandhi, and Mitesh M. Khapra. 2025. Rasmalai : Resources for Adaptive Speech Modeling in IndiAn Languages with Accents and Intonations. In *Interspeech 2025*, pages 4128–4132.

Matthew Snover, Bonnie Dorr, Rich Schwartz, Linnea Micciulla, and John Makhoul. 2006. A study of translation edit rate with targeted human annotation. In *Proceedings of the 7th Conference of the Association* *for Machine Translation in the Americas: Technical Papers*, pages 223–231, Cambridge, Massachusetts, USA. Association for Machine Translation in the Americas.

Yixuan Su, Tian Lan, Yan Wang, Dani Yogatama, Lingpeng Kong, and Nigel Collier. 2022. A contrastive framework for neural text generation. *Preprint*, arXiv:2202.06417.

Rohan Taori, Ishaan Gulrajani, Tianyi Zhang, Yann Dubois, Xuechen Li, Carlos Guestrin, Percy Liang, and Tatsunori B. Hashimoto. 2023. Stanford alpaca: An instruction-following llama model. `https://github.com/tatsu-lab/stanford_alpaca`.

Qwen Team. 2024. Qwen2.5: A party of foundation models.

Ashish Vaswani, Noam Shazeer, Niki Parmar, Jakob Uszkoreit, Llion Jones, Aidan N Gomez, Ł ukasz Kaiser, and Illia Polosukhin. 2017. Attention is all you need. In *Advances in Neural Information Processing Systems*, volume 30. Curran Associates, Inc.

Koki Wataoka, Tsubasa Takahashi, and Ryokan Ri. 2025. Self-preference bias in llm-as-a-judge. *Preprint*, arXiv:2410.21819.

Vicky Zayats, Mari Ostendorf, and Hannaneh Hajishirzi. 2016. Disfluency detection using a bidirectional lstm. *Preprint*, arXiv:1604.03209.

Lianmin Zheng, Wei-Lin Chiang, Ying Sheng, Siyuan Zhuang, Zhanghao Wu, Yonghao Zhuang, Zi Lin, Zhuohan Li, Dacheng Li, Eric P. Xing, Hao Zhang, Joseph E. Gonzalez, and Ion Stoica. 2023. Judging llm-as-a-judge with mt-bench and chatbot arena. *Preprint*, arXiv:2306.05685.

## A ASR Behavior and Disfluency Preservation

A core motivation for our work is that real ASR transcripts of spontaneous speech often contain fillers, repetitions, false starts, and self-repairs. While recent ASR systems-including those incorporating large language models-aim to produce cleaner transcripts, it remains unclear whether such systems reliably remove conversational disfluencies in practice. We empirically examine the behavior of the ASR front-ends used in our pipeline and show that, for Indic conversational speech, disfluencies are not consistently normalized away and instead remain prevalent in the ASR output. Consequently, disfluency phenomena are frequently preserved rather than systematically removed by current state-of-the-art ASR systems.

**ASR front-ends.** Our transcripts are generated using two strong ASR models: (i) Whisper v3 Large, selected for its multilingual robustness under noisy conditions, and (ii) AI4Bharat Indic Conformer (600M), trained extensively on diverse Indian-language corpora and designed to capture Indic phonetics and accent variability. While these systems may perform standard text normalization (e.g., orthographic consistency), they do *not* consistently normalize away conversational disfluencies.

**What disfluencies are preserved?** Across Bengali, Hindi, and Marathi, we observe a rich spectrum of disfluencies in ASR output, including: filled pauses (e.g., তো [toh], মানে [maane], हाँ [haan], अ [a]), whole-word and phrase repetitions (e.g., repeated multi-word fragments), partial-word stutters and broken syllables (e.g., ক্ষে-ক্ষে-ক্ষে... [kshe-kshe-kshe…], स...ऊबर... [s…ubar…], ना प्रति ना प्रति... [na prati na prati…]), false starts and abandoned clauses (e.g., हो प्र... [ho pr…], अरे— [are—]), and self-repairs where speakers restart or correct themselves mid-utterance (e.g., জামি → আমি [jami → aami], कोई उसके हाँ → उसके गाँव में [koi uske haan → uske gaon mein], मी जाड जाते → वेट कमी केला [mi jaad jaate → vet kami kela]). We also observe non-speech noise and garbled segments typical of spontaneous speech and challenging acoustics.

**Expert validation and prevalence in real audio.** All disfluency patterns above were reviewed and validated by language and speech-processing experts using transcripts produced by Whisper v3 Large [1] and the AI4Bharat Indic Conformer[2]. Importantly, the underlying audio is drawn from real-world conversational YouTube recordings (podcasts/interviews) rather than scripted speech; in this setting, approximately 30% of sentences contain at least one detectable disfluency. This confirms that disfluencies are a systematic property of the data distribution encountered in practice, not an artifact of annotation.

**LLM-powered ASR is not a reliable counterexample for Indic speech.** For completeness, we evaluated an LLM-based audio model Qwen/Qwen2-Audio-7B [3] on Bengali, Hindi, and Marathi. In our experiments, it failed to produce meaningful ASR outputs across all three languages: even with language-specific in-context demonstrations, the model *ignored the audio* and echoed the demonstrations verbatim, indicating insufficient acoustic–phonetic grounding for Indic speech. Representative qualitative examples are shown in Figure 3.

**Implication for our pipeline.** These observations establish the gap motivating our work: even with strong ASR front-ends, *natural conversational disfluencies frequently remain in transcripts* and cannot be assumed to be removed by upstream ASR. Therefore, a dedicated disfluency-correction module remains important for downstream LLM-based systems and speech applications operating on real ASR output.

## B Impact of Disfluencies on Downstream LLM Tasks

This work is motivated by the hypothesis that disfluent transcripts can degrade downstream LLM performance by introducing spurious tokens, broken syntax, and inconsistent discourse structure. To verify this hypothesis empirically (beyond reference-based cleaning metrics), we evaluate modern instruction-tuned LLMs on two downstream tasks that explicitly stress comprehension and reasoning: Question Answering (QA) and Machine Translation (MT). In both tasks, we compare performance under fluent versus disfluent context derived from real ASR-style inputs.

[1] https://huggingface.co/openai/whisper-large-v3

[2] https://huggingface.co/ai4bharat/indic-conformer-600m-multilingual

[3] https://huggingface.co/Qwen/Qwen2-Audio-7B

| Language | Qwen2-Audio-7B Output | Ground Truth |
|---|---|---|
| Bengali | আমি বাড়ি যাব | তো আমি মানে কাল রাতে ভাবছিলাম, উhm আবার কি শুরু করবো নাকি… |
| | এটা ঠিক না | এটা উh ঠিক বুঝতে পারছি না, তুমি আবার একবার বোলো তো… |
| Hindi | मैं अच्छा हूँ | अच्छा, वो… मतलब मैं कल ही सोच रहा था कि क्या करना है आगे। |
| | हाँ ज़रूर, मैं पाँच बजे पहुँ | ये उhm थोड़ा अटक रहा है, तुम एक बार फिर से बता दो न… |
| Marathi | माझ्या राजा, तू चांगला माणूस आहेस. | अं, म्हणजे काल मी विचार करत होतो की पुढे काय करायचं… |
| | उद्या माझ्या घरी या. | हे उhm जरा नीट ऐकू येत नाहीये, तू पुन्हा एकदा सांगशील का… |

Figure 3: Qualitative comparison of ASR behavior on disfluent speech across Hindi, Bengali, and Marathi. Qwen2-Audio-7B outputs are contrasted with ground-truth transcripts.

| Language | Model | Avg. Score (Disfluent) | Avg. Score (Fluent) |
|---|---|---|---|
| Hindi | LLaMA | 1.18 | 1.70 |
| | Qwen | 2.14 | 2.80 |
| Marathi | LLaMA | 1.12 | 1.58 |
| | Qwen | 1.74 | 2.16 |
| Bengali | LLaMA | 2.20 | 2.76 |
| | Qwen | 1.96 | 2.62 |

Table 6: Average human evaluation scores (1–5 scale) for QA robustness across 100 manually constructed question–answer pairs per language. Higher scores indicate better answer quality.

### B.1 Human-rated QA under fluent vs. disfluent context.

We create 100 QA instances per language (Hindi, Marathi, Bengali) by sampling paired fluent and disfluent sentences and asking native-language experts to author questions and evaluate model answers. Specifically, experts rate answer quality on a 1–5 Likert scale, where higher scores indicate more accurate and context-faithful responses. Results are reported in Table 6.

Across all languages and both backbones Llama-3.2-3B-Instruct and Qwen2.5-3B-Instruct, disfluent context consistently reduces answer quality. For Hindi, LLaMA drops from 1.70 (fluent) to 1.18 (disfluent), and Qwen drops from 2.80 to 2.14. Similar degradations appear for Marathi (1.58 $\rightarrow$ 1.12 for LLaMA; 2.16 $\rightarrow$ 1.74 for Qwen) and Bengali (2.76 $\rightarrow$ 2.20 for LLaMA; 2.62 $\rightarrow$ 1.96 for Qwen). These consistent declines indicate that even strong instruction-tuned LLMs are not fully robust to ASR-style disfluencies when required to *interpret* and *reason* over noisy conversational context.

### B.2 MT robustness on real ASR data under fluent vs. disfluent input.

We additionally test robustness in a real deployment-like MT setting using 100 randomly sampled *real-ASR* sentences, which naturally include noisy segmentation, inconsistent pauses, and conversational artifacts. For each sample, native-language experts provide gold translations. We then compare MT outputs generated from fluent versus disfluent versions of the same utterance. Results are summarized in Table 7.

| Language Pair | Model | ΔBLEU (Fl–Dis) | ΔchrF2 (Fl–Dis) | ΔTER (Dis–Fl) |
|---|---|---|---|---|
| Hindi → Bengali | Qwen | +3.9 | +1.9 | +8.1 |
| | Llama | +2.0 | +0.8 | +3.9 |
| Marathi → Hindi | Qwen | +4.3 | +2.2 | +12.5 |
| | Llama | +3.5 | +1.8 | +13.1 |
| Bengali → Hindi | Qwen | +4.4 | +1.8 | +17.4 |
| | Llama | +4.7 | +2.2 | +17 |

Table 7: Difference in translation quality between **fluent** and **disfluent** inputs for Qwen and Llama. Positive values across BLEU and chrF2 indicate higher scores for fluent sentences. Positive TER differences (Disfluent – Fluent) show that disfluent inputs require substantially more edits, confirming that LLMs are not robust to disfluencies across all language pairs.

Disfluent inputs cause substantial and consistent degradation in translation quality across language pairs. For Hindi→Bengali, Qwen's BLEU decreases by 3.9 points when translating disfluent versus fluent input, while LLaMA decreases by 2.0 points. For Marathi→Hindi, the BLEU drops are 4.3 (Qwen) and 3.5 (LLaMA), and for Bengali→Hindi they are 4.4 (Qwen) and 4.7 (LLaMA). In all cases, TER increases sharply (by approximately +8 to +17 points), indicating that substantially more edits are required to match the reference translation. These results show that disfluencies do not merely introduce superficial noise: they lead to misinterpretations and mistranslations even in modern LLM-based MT pipelines.

### B.3 Takeaway

Together, the QA and MT evaluation provide converging evidence that contemporary instruction-tuned LLMs are *not* inherently robust to natural ASR disfluencies. Performance drops are consistent across languages, models, and tasks that require understanding rather than surface-level rewriting. This directly supports the continued relevance of disfluency correction and motivates our contrastive training framework, which is designed to suppress disfluent content while preserving meaning in real-world ASR transcripts.

## C Task Relevance and Application Scenarios

Disfluency correction is relevant whenever downstream systems consume ASR transcripts produced from spontaneous speech. Common deployment settings include customer-support call centers, voice assistants, live captioning, and accessibility tools, where the ASR output is subsequently used for (i) semantic understanding and decision making (e.g., LLM-based chat/QA), (ii) translation (e.g., cross-lingual support), and (iii) speech generation (e.g., text-to-speech for agent responses or spoken summaries). In these pipelines, fillers, repetitions, false starts, and self-repairs can introduce spurious tokens and unstable phrasing that degrade both text understanding and speech rendering.

### C.1 Text-based downstream tasks.

Appendix B shows that disfluencies harm LLM performance on QA and MT under fluent vs. disfluent comparisons. These degradations directly impact LLM-powered voice interfaces and conversational agents, where a system must interpret user speech accurately and respond coherently under realistic ASR noise.

### C.2 TTS evaluation: disfluencies degrade speech naturalness and prosody.

In addition to QA and MT, we evaluate the impact of disfluencies on Text-to-Speech (TTS), since many voice pipelines synthesize speech from ASR transcripts (e.g., spoken translations, agent responses, or readbacks of recognized text). We conduct a human Mean Opinion Score (MOS) study on 25 randomly sampled disfluent ASR sentences per language (Hindi, Marathi, Bengali). Native evaluators rate synthesized speech on a 1–5 scale along multiple perceptual dimensions: *Naturalness*, *Pronunciation Accuracy*, *Prosody*, *Pausing & Timing*, *Emotional Expression*, and *Overall Audio Quality*.

Across languages and TTS models, disfluent transcripts consistently yield low MOS and perceptibly degraded speech quality. For Hindi, `ai4bharat/indic-parler-tts` (Sankar et al., 2025; Lacombe et al., 2024; Lyth and King, 2024)[4] and suno/bark[5] achieve average MOS scores of 2.09 and 2.03, respectively, on disfluent inputs. For Marathi, disfluent inputs yield MOS scores of 2.27 with indic-parler-tts and 2.09 with facebook/mms-tts-mar[6] (Pratap et al., 2023). For Bengali, disfluent inputs yield MOS scores of 2.46 with indic-parler-tts and 2.40 with facebook/mms-tts-ben[7] (Pratap et al., 2023). Evaluators consistently report that disfluencies induce unnatural pauses, broken rhythm, and erratic prosody; in contrast, fluent rewrites produce noticeably smoother timing and more natural-sounding speech.

### C.3 Why disfluency correction does not contradict naturalness.

A potential concern is that removing disfluencies might make speech sound less natural by eliminating conversational cues. In our setting, the goal is not to erase all stylistic variability, but to remove *ASR-induced* and *speech-disfluency artifacts* (e.g., repeated fragments, abandoned clauses, filled pauses) that are rendered unnaturally by TTS systems. The MOS results above indicate that, for current Indic TTS models, preserving such artifacts in the text typically harms naturalness and prosody rather than improving conversational realism. Disfluency correction therefore serves as a practical pre-processing step for producing intelligible, well-timed synthesized speech from real ASR transcripts.

### C.4 Takeaway.

Taken together, the QA and MT results (Appendix B) and the MOS findings here show that disfluencies negatively affect multiple downstream components in ASR-driven pipelines, spanning reasoning, translation, and speech synthesis. This motivates a dedicated disfluency-correction module in real-world LLM-powered voice interfaces and supports our focus on learning objectives that suppress disfluent content while preserving meaning.

[4] https://huggingface.co/ai4bharat/indic-parler-tts
[5] https://huggingface.co/suno/bark
[6] https://huggingface.co/facebook/mms-tts-mar
[7] https://huggingface.co/facebook/mms-tts-ben

| Train→Test | Dataset | BLEU↑ | chrF2↑ | TER↓ |
|---|---|---|---|---|
| Hindi→Hindi | Manual | 92.4 | 96.5 | 8.9 |
| | Real | 69.8 | 91.2 | 31.2 |
| Hindi→Bengali | Manual | 87.1 | 92.3 | 13.7 |
| | Real | 64.9 | 88.2 | 26.8 |
| Hindi→Marathi | Manual | 83.4 | 96.4 | 14.2 |
| | Real | 75.7 | 93.6 | 17.9 |
| Marathi→Marathi | Manual | 88.8 | 97.7 | 10.4 |
| | Real | 75.7 | 94.5 | 19.6 |
| Marathi→Bengali | Manual | 86.4 | 91.3 | 15.0 |
| | Real | 64.0 | 87.3 | 28.4 |
| Marathi→Hindi | Manual | 90.1 | 97.2 | 8.0 |
| | Real | 82.2 | 94.3 | 12.9 |
| Bengali→Bengali | Manual | 90.9 | 93.6 | 10.1 |
| | Real | 71.5 | 89.3 | 21.4 |
| Bengali→Hindi | Manual | 91.1 | 97.2 | 6.4 |
| | Real | 77.9 | 91.4 | 12.1 |
| Bengali→Marathi | Manual | 90.3 | 98.0 | 6.7 |
| | Real | 81.2 | 95.5 | 10.2 |

Table 8: Zero-shot cross-lingual evaluation of LLaMA models. Each model is fine-tuned on a single language and evaluated on the same or different languages without additional supervision, illustrating cross-lingual transfer.

## D Multilingual vs. Language-Specific Instruction Fine-tuning

We investigate: (1) multilingual versus language-specific fine-tuning performance, and (2) zero-shot cross-lingual transfer across related Indic languages. All results are reported on the disfluency correction task using the same evaluation protocol as in the main paper (BLEU/chrF2↑, TER↓).

### D.1 Setup

We fine-tune Llama-3.2-3B-Instruct under two settings: (1) Multilingual FT, trained jointly on Hindi, Bengali, and Marathi; and (2) Language-Specific FT, where separate models are fine-tuned per language. To study cross-lingual transfer, we evaluate each language-specific model on the other languages without additional supervision. All runs use identical hyperparameters and data preprocessing.

### D.2 Results

**Multilingual vs. Language-Specific (manually edited).** On manually edited data, multilingual FT yields stronger in-language performance overall (Table 3). In particular, multilingual FT improves BLEU over language-specific FT by +0.1 (Hindi), +3.3 (Bengali), and +5.6 (Marathi), suggesting that joint training provides beneficial sharing across languages in the clean-text setting.

**Multilingual vs. Language-Specific (real ASR).** On real ASR data, the comparison is more mixed

| Language | Dataset | Qwen2.5-3B-Instruct | | | Llama-3.2-3B-Instruct | | |
|---|---|---|---|---|---|---|---|
| | | BLEU ↑ | chrF2 ↑ | TER ↓ | BLEU ↑ | chrF2 ↑ | TER ↓ |
| Hindi | Manually Edited | **77.5** | **87.8** | **17.1** | 70.1 | 82.5 | 26.1 |
| | Real Data | **56.0** | **75.4** | **28.1** | 38.6 | 62.4 | 40.5 |
| Bengali | Manually Edited | **73.1** | **89.4** | **18.1** | 66.4 | 84.9 | 24.3 |
| | Real Data | **56.8** | **83.2** | **28.5** | 49.1 | 74.3 | 40.3 |
| Marathi | Manually Edited | 69.8 | **87.3** | 21.1 | **71.2** | 85.6 | **24.4** |
| | Real Data | 59.2 | **84.0** | 26.5 | **63.1** | 81.1 | **28.9** |

Table 9: Comparison of Qwen2.5-3B-Instruct and Llama-3.2-3B-Instruct (zero-shot, without finetuning) on manually edited and real datasets across Hindi, Bengali, and Marathi. Best scores per metric are bolded.

(Table 8). Language-specific FT improves BLEU for Hindi (+5.0) and Bengali (+1.9) relative to multilingual FT and often reduces TER by a few points, indicating better robustness to language- and domain-specific ASR artifacts. Marathi, however, shows a drop in BLEU under language-specific FT relative to multilingual FT, suggesting that multilingual training can still provide beneficial regularization for some languages under noisy conditions.

**Zero-shot cross-lingual transfer.** We observe substantial cross-lingual generalization among these languages (Table 8). For example, Hindi-only FT achieves 87.1 BLEU on Bengali and 83.4 BLEU on Marathi (manually edited), while Marathi-only FT achieves 90.1 BLEU on Hindi and 86.4 BLEU on Bengali. Bengali-only FT also transfers strongly to Hindi (91.1 BLEU) and Marathi (90.3 BLEU). On real ASR data, transfer persists with expected degradation, but performance remains high (typically mid-60s to low-80s BLEU depending on the direction), indicating that a single-language finetune can still provide meaningful improvements for related languages.

### D.3 Takeaway

Multilingual FT is generally preferable on clean manually edited data, while language-specific FT can be advantageous under real ASR noise for some languages (notably Hindi and Bengali). Across all settings, we observe strong zero-shot transfer among the studied languages, supporting the use of joint or low-cost single-language fine-tuning when full multilingual supervision is unavailable.

## E Zero-shot Prompting on base LLMs

Recent studies, such as (Lima and Campelo, 2024) have explored zero-shot prompting for English speech transcripts, showing that LLMs can often identify and remove disfluent spans without explicit supervision. However, English disfluencies are relatively simpler: less morphologically entangled and syntactically constrained than those observed in Indian languages. In Indic contexts, disfluency correction becomes substantially more complex due to features, such as rich case marking, verb agreement, and compound verb constructions. This linguistic diversity motivates a systematic investigation into whether multilingual LLMs can extend such implicit capabilities to morphologically rich, low-resource settings without any task-specific finetuning.

To this end, we evaluate Qwen2.5-3B-Instruct and Llama-3.2-3B-Instruct in a zero-shot prompting setup, where models receive only a natural-language instruction and a disfluent input, with no gradient updates or supervised examples. Across all three languages, Qwen consistently performs better than Llama in BLEU, chrF2, and TER, reflecting stronger multilingual grounding and transfer. From Table 9, Qwen attains 77.5 BLEU, 87.8 chrF2 and 17.1 TER on the *Hindi–Manually Edited* subset, whereas Llama reaches 70.1, 82.5, and 26.1, respectively. Nevertheless, both models show substantial degradation on *Real Data*, indicating that even robust instruction-tuned LLMs struggle with spontaneous and morphologically complex disfluencies unless explicitly exposed to such noise during training.

Overall, these results suggest that while zero-shot prompting may suffice for relatively analytic languages such as English, it remains inadequate for Indic languages. The morphological and syntactic richness of these languages necessitates explicit disfluency-aware fine-tuning or contrastive instruction-based training to achieve fluent and faithful corrections.

## F Experimental Setup

We fine-tuned the Llama-3.2-3B-Instruct model using an instruction-tuning setup optimized with a contrastive loss. The model was trained with a sequence length of 512 tokens and an effective batch size of 16, achieved through gradient accumulation of 2 steps with a per-device batch size of 8. The learning rate was set to $1e^{-5}$ and followed a cosine decay schedule with a warmup ratio of 0.1. A contrastive weight $\lambda = 0.3$ was applied, with the same fraction used for warmup. We used label smoothing of 0.01 to improve generalization and trained in bfloat16 precision with gradient checkpointing enabled to reduce memory usage. Optimization was performed using the fused AdamW optimizer, and early stopping was employed with a patience of 3 epochs. All experiments were conducted on a single NVIDIA A100 GPU (80 GB). We use publicly available models specifically Qwen2.5-3B-Instruct[8] is available with Qwen Research License Agreement. Llama-3.2-3B-Instruct[9] is available under Llama 3.2 Community License. MuRIL[10] is available under Apache license. We used generative AI tool - specifically ChatGPT (OpenAI et al., 2024) for text editing and grammatical correction. We used single runs for all our experiments.

## G Example Outputs

### G.1 Hindi Example

We show a Hindi example in Figure 4 (Row 1) exhibiting repeated fillers and a false start.

**Baseline behavior:** Multilingual Instruction removes some noise but leaves partial fillers अ म यह अ (uh, I, this, uh), showing under-deletion. mBART drops the contrastive connective पर (but), weakening discourse structure and pragmatic contrast. W/o Contrastive retains पर but over-compresses the clause पर दोहरा जीवन मुश्किल होता है (but a double life becomes difficult), losing the natural rhythm of the actual sentence (but those who lead a double life, they find it difficult.)

**Our approach:** Our contrastive instruction-tuned model produces: पर ये जो दोहरा जीवन होता है यह मुश्किल होता है। It fully removes hesitation tokens while preserving (i) the discourse marker पर (but), (ii) the clauses ये जो and यह होता है. This balance of deletion and retention maintains both adequacy and fluency. Here, the MuRIL-based token-level supervision is combined with an *anti-disfluency contrastive loss* that explicitly penalizes regeneration of disfluent spans. This enables precise noise removal while maintaining semantic relation with the source.

### G.2 Bengali Example

Figure 4 (Row 2) illustrates a Bengali utterance with fillers, repetitions, and false restarts.

**Baseline behavior:** Multilingual Instruction and mBART retain multiple discourse-address terms and subject restarts স্যার ... আমি ... আমি (sir ... I ... I), leaving residual repetition একটু খানি ... একটু বেশি (a little bit ... a bit more) and occasional pronoun stacking আমরা আমি (we I). W/o Contrastive exacerbates this with compounded subjects and self-repairs, reflecting weak boundary control over disfluent spans.

**Our approach:** Our contrastive instruction-tuned model outputs: আমি এই বিষয়ে বলতে চাই স্যার ২০১৫ এর আগে আমাদের স্কুলে এনসিসি হত তো ডিফেন্স সম্বন্ধে আমার একটু বেশি চাহিদা ছিল (I want to speak on this topic, sir - before 2015, our school had NCC, so I had a bit more interest in defense).
It (i) removes the filler আঃ (uh),
(ii) collapses redundant address/subject restarts স্যার ... আমি ... আমি (sir ... I ... I) $\rightarrow$ single স্যার (sir) and single আমি (I),
(iii) eliminates pronoun stacking আমরা আমি (we I).

The result is fluent, yet faithful to the source utterance. Compared to baselines, our model achieves *clean deletion with structure preservation* (single address term, single subject, intact temporal–context–preference flow).

### G.3 Marathi Example

Figure 4 (Row 3) presents a Marathi example containing partial word repetition and redundant discourse markers.

**Baseline behavior:** Multilingual Instruction and mBART both retain partial reduplication सूच सूचना (instruc– instruction) and an unnecessary connective की किंवा सांग (that, or, say), indicating model's inability to remove mistaken words when the speaker self-corrects the same. W/o Contrastive deletes portions of the repetition but distorts morphology नवऱ्याला (to the husband; truncated as husb–) and omits the complement marker की (that), generates a slightly ungrammatical phrasing.

[8] https://huggingface.co/Qwen/Qwen2.5-3B-Instruct
[9] https://huggingface.co/meta-llama/Llama-3.2-3B-Instruct
[10] https://huggingface.co/google/muril-base-cased

| **Disfluent Sentences** | **Model Outputs** |
|---|---|
| **Hindi :** अ अ म पर ये जो दोहरा जीवन होता है यह अ अ मुश्किल होता है। | **Multilingual Instruction:** अ म पर ये जो दोहरा जीवन होता है यह अ मुश्किल होता है।<br>**mBART :** ये जो दोहरा जीवन होता है मुश्किल होता है।<br>**W/o Contrastive :** पर दोहरा जीवन मुश्किल होता है।<br>**Our approach :** पर ये जो दोहरा जीवन होता है यह मुश्किल होता है। |
| **Bengali :** আঃ স্যার আমি এই বিষয়ে বলতে চাই স্যার আমি ২০১৫ এর আগে আমি আমাদের স্কুলে এনসিসি হত তো আমরা আমি ডিফেন্স সম্বন্ধে একটু খানি আমার একটু বেশি চাহিদা ছিল | **Multilingual Instruction:** স্যার আমি এই বিষয়ে বলতে চাই স্যার আমি ২০১৫ এর আগে আমি আমাদের স্কুলে এনসিসি হত তো আমি ডিফেন্স সম্পর্কে একটু খানি আমার একটু বেশি চাহিদা ছিল<br>**mBART :** স্যার আমি এই বিষয়ে বলতে চাই স্যার আমি ২০১৫ এর আগে আমাদের স্কুলে এনসিসি হত তো আমি ডিফেন্স সম্বন্ধে একটু খানি আমার একটু বেশি চাহিদা ছিল<br>**W/o Contrastive :** স্যার আমি এই বিষয়ে বলতে চাই স্যার আমি ২০১৫ এর আগে আমি আমাদের স্কুলে এনসিসি হত তো আমরা আমি ডিফেন্স সম্বন্ধে একটু খানি আমার একটু বেশি চাহিদা ছিল<br>**Our approach :** আমি এই বিষয়ে বলতে চাই স্যার ২০১৫ এর আগে আমাদের স্কুলে এনসিসি হত তো ডিফেন্স সম্বন্ধে আমার একটু বেশি চাহিদা ছিল |
| **Marathi :** सर मी त्या युवतीला आणि तिच्या प्र पहिल्या नवऱ्याला दोघांना बोलवून घेईन आणि दोघांना सूच सूचना करेन की किंवा सांग समजावण्याचा पहिला प्रयत्न करेन | **Multilingual Instruction:** सर मी त्या युवतीला आणि तिच्या पहिल्या नवऱ्याला दोघांना बोलवून घेईन आणि दोघांना सूचना करेन की किंवा सांग समजावण्याचा पहिला प्रयत्न करेन<br>**mBART :** सर मी त्या युवतीला आणि तिच्या पहिल्या नवऱ्याला दोघांना बोलवून घेईन आणि दोघांना सूच सूचना करेन की किंवा सांग समजावण्याचा पहिला प्रयत्न करेन<br>**W/o Contrastive :** सर मी त्या युवतीला आणि तिच्या पहिल्या नवयाला दोघांना बोलवून घेईन आणि दोघांना सूच करेन की किंवा समजावण्याचा पहिला प्रयत्न करेन<br>**Our approach :**सर मी त्या युवतीला आणि तिच्या पहिल्या नवऱ्याला दोघांना बोलवून घेईन आणि दोघांना सूचना करेन किंवा समजावण्याचा पहिला प्रयत्न करेन |

Figure 4: Comparison of disfluent sentences and model outputs across Hindi, Bengali, and Marathi. Our Approach consistently produces fluent outputs closely matching references compared to baselines.

**Our approach:** Our contrastive instruction-tuned model outputs: सर मी त्या युवतीला आणि तिच्या पहिल्या नवऱ्याला दोघांना बोलवून घेईन आणि दोघांना सूचना करेन किंवा समजावण्याचा पहिला प्रयत्न करेन. (Sir, I will call that young woman and her first husband, give both of them instructions or make the first attempt to explain.) It precisely removes redundant fragments सूच (instruc–), सांग (say)) while preserving the syntactic connector किंवा (or) and the alignment of subordinate clauses. The output is grammatically correct, semantically faithful, and rhythmically natural, reflecting complete fluency restoration.

## H Human Evaluation Analysis

To complement the automatic metrics and LLM-as-a-Judge evaluation, we conducted a human evaluation to assess disfluency correction quality in terms of fluency and meaning preservation. For each language (Hindi, Bengali, and Marathi), we randomly sampled 100 sentences from the evaluation set. Native-language experts rated system outputs on a 5-point Likert scale (1 = very poor, 5 = excellent), jointly considering grammatical fluency and faithfulness to the original meaning. We provide the detailed instructions in Appendix J.

Across all three languages, the proposed contrastive learning approach consistently outperforms multilingual instruction fine-tuning using LLaMA as the underlying model. For Hindi, the contrastive model achieves a mean human score of 4.53 compared to 3.89 for multilingual instruction fine-tuning, corresponding to an improvement of +0.64. In addition, 91% of Hindi outputs produced by the contrastive model receive high-quality ratings (score $\geq 4$), compared to 81% for the baseline.

For Bengali, the contrastive approach improves the mean human score from 3.93 to 4.20 (+0.27), and increases the proportion of high-quality outputs from 64% to 77%. For Marathi, which exhibits relatively fewer disfluencies, the contrastive model still yields consistent gains, improving the mean score from 4.69 to 4.81 (+0.12) and increasing the proportion of outputs rated $\geq 4$ from 93% to 98%.

| Language | Dataset | Zero-Shot | | | Few-Shot | | |
|---|---|---|---|---|---|---|---|
| | | BLEU↑ | chrF2↑ | TER↓ | BLEU↑ | chrF2↑ | TER↓ |
| Hindi | Manually Edited | 94.3 | 97.3 | 4.6 | 94.9 | 98.1 | 3.7 |
| | Real Data | 82.5 | 91.6 | 9.6 | 87.2 | 93.6 | 6.8 |
| Bengali | Manually Edited | 93.4 | 97.7 | 4.8 | 94.3 | 98.5 | 4.0 |
| | Real Data | 79.7 | 92.1 | 12.4 | 79.6 | 91.8 | 12.5 |
| Marathi | Manually Edited | 92.7 | 96.9 | 5.7 | 93.9 | 97.7 | 4.7 |
| | Real Data | 87.7 | 94.8 | 6.9 | 86.9 | 94.5 | 8.3 |

Table 10: GPT-4o performance under zero-shot and few-shot (2-shot) prompting on manually edited and real ASR test sets across Hindi, Bengali, and Marathi.

These trends closely mirror the findings from the automatic metrics and the LLM-as-a-Judge evaluation, providing converging evidence that contrastive supervision leads to perceptible improvements in real-world correction quality. Notably, the largest gains are observed for Hindi, which contains greater disfluency variability in conversational ASR transcripts, suggesting that contrastive learning is particularly effective under noisier conditions.

Qualitative inspection of system outputs further reveals that the contrastive model is especially effective at correcting common disfluency phenomena, including repetitions, fillers, discourse markers, false starts, and mid-sentence self-repairs. These behaviors align closely with the disfluency categories defined in our annotation guidelines, which are released alongside our code repository. Overall, the human evaluation confirms that the proposed contrastive learning framework produces more fluent and semantically faithful corrections in practice.

| Language | BLEU↑ | chrF2↑ | TER↓ |
|---|---|---|---|
| Hindi | 70.5 | 85.1 | 16.6 |
| Bengali | 61.8 | 81.1 | 21.3 |
| Marathi | 74.9 | 89.6 | 12.7 |

Table 11: Gemini 2.5 Pro zero-shot performance on real ASR test data across Hindi, Bengali, and Marathi.

## I Comparison with Frontier Proprietary Models

A natural question is whether the gains achieved by our contrastive pipeline could be replicated by simply prompting a much larger proprietary frontier model with a well-designed instruction. To investigate this, we evaluate two of the most capable frontier models - GPT-4o and Gemini 2.5 Pro - on the disfluency correction task across Hindi, Bengali, and Marathi. For GPT-4o, we consider three prompting strategies: zero-shot, where the model is directly instructed to remove disfluencies without any examples; few-shot (2-shot), where two parallel disfluent–fluent sentence pairs are provided as in-context demonstrations; and zero-shot chain-of-thought (CoT), where the model is instructed to first identify disfluent spans and then reconstruct the fluent sentence before producing the final output. For Gemini 2.5 Pro, due to API cost and rate limit constraints, we report zero-shot results on the real test set only. All evaluations use the same metrics (BLEU, chrF2, TER) and test sets as the main paper, with GPT-4o results reported in Table 10 and Gemini 2.5 Pro results in Table 11.

Zero-shot CoT prompting yields results comparable to few-shot across all three languages and for both the evaluation sets, with no consistent improvement over standard few-shot prompting, confirming that explicit step-by-step reasoning does not offer additional gains for this task.

From Table 3 and Table 4), our Qwen-2.5-3B contrastive model outperforms GPT-4o few-shot (its strongest setting) on 4 out of 6 conditions on BLEU and 4 out of 6 on chrF2, with simultaneous gains on both metrics in 3 out of 6 conditions, despite being significantly smaller. Specifically, our Qwen outperforms GPT-4o few-shot by +3.9 BLEU on Hindi real data and +2.1 BLEU on Bengali manually edited data. In the two settings where GPT-4o leads on BLEU, the margins are small and do not hold consistently across chrF2. Against Gemini 2.5 Pro, our contrastive model outperforms by substantial margins on all three languages: +20.6 BLEU and +9.8 chrF2 on Hindi, +14.1 BLEU and +12.1

chrF2 on Bengali, and +9.5 BLEU and +4.7 chrF2 on Marathi (all on real data). Taken together, these results demonstrate that neither scaling to a frontier model nor applying advanced prompting strategies consistently closes the gap with task-specific contrastive training. The morphological richness and syntactic complexity of Indic disfluency patterns - involving complex word formations, verb-final constructions, and code-mixed repairs - require explicit disfluency-aware supervision that zero-shot or few-shot prompting of even the strongest available models cannot substitute.

## J Human Rating Guidelines

You are given two sentences: (1) a fluent reference sentence (gold), and (2) a model-generated output. Your task is to rate the model-generated output on a scale from 1 to 5, based only on fluency and naturalness. Do not judge factual correctness or meaning accuracy. Even if the sentence is factually incorrect, it can still receive a high score if it sounds fluent and natural.

**What to focus on**

- Whether the sentence sounds natural to a native speaker
- Whether it flows smoothly without awkward pauses
- Whether it is free from disfluencies such as repetitions, fillers, or broken structure

**Common disfluency markers**

- Repetitions (words or phrases repeated unnecessarily)
- Fillers or discourse markers (e.g., *matlab*, *mane*, ওই, म्हणजे)
- False starts and self-corrections
- Truncated or incomplete words
- Elongated sounds or hesitation pauses

**Scoring guidelines**

- **1 – Completely disfluent:** Very hard to understand, with severe repetition, broken structure, and unclear meaning.
- **2 – Highly disfluent:** Frequent fillers, repetitions, or false starts. Meaning is only partially understandable.
- **3 – Moderately disfluent:** Core meaning is understandable, but noticeable disfluencies remain.
- **4 – Mostly fluent:** Minor hesitations or fillers that do not affect understanding.
- **5 – Fully fluent:** Completely natural and smooth, with no disfluencies.

## K Instruction Prompt Used

The following instruction was used as the system prompt for our proposed contrastive disfluency correction model:

```
You are given a disfluent sentence generated
    by an Automatic Speech Recognition (ASR)
    system.
The sentence may contain disfluencies such as
    repetitions, fillers (e.g., 'um', 'uh'),
    discourse
markers (e.g., 'you know', 'I mean'), or false
     starts in {Language}.

Your task is to remove these disfluencies
    while preserving the original meaning and
    grammatical correctness.

You are also provided with:
- The disfluent sentence
- A tokenized version of the sentence
- A sequence of predicted labels for each
    token, where:

   * '1' = the token is disfluent and should
       be removed

   * '0' = the token is fluent and should be
       retained
- A list of disfluent tokens that must be
    removed from the sentence

Using this information, reconstruct the fluent
     sentence.
Make sure to remove all tokens listed as
    disfluent while preserving meaning and
    grammatical correctness.

Tokenized Input: {tokens}
Predicted Labels: {labels}
Disfluent Sentence: {disfluent}
Disfluent Tokens: {disfluent_tokens}
Fluent Sentence:
```